# On the variants of SVM methods applied to GPR data to classify tack coat characteristics in French pavements: two experimental case studies

Grégory Andreoli
AAn/ENDSUM, Angers
Cerema – France
gregory.andreoli@cerema.fr

Amine Ihamouten
MAST/LAMES, Nantes
Gustave Eiffel University – France
amine.ihamouten@univ-eiffel.fr

Mai Lan Nguyen
MAST/LAMES, Nantes
Gustave Eiffel University – France
mai-lan.nguyen@univ-eiffel.fr

Yannick Fargier
GERS/RRO, Bron
Gustave Eiffel University – France
yannick.fargier@univ-eiffel.fr

Cyrille Fauchard
DGI/ENDSUM, Rouen
Cerema – France
cyrille.fauchard@cerema.fr

Jean-Michel Simonin
MAST/LAMES, Nantes
Gustave Eiffel University – France
jean-michel.simonin@univ-eiffel.fr

Viktoriia Buliuk
GERS/GeoEND, Nantes
Gustave Eiffel University – France
viktoriia.buliuk@univ-eiffel.fr

David Souriou
MAST/LAMES, Nantes – FI-NDT
Gustave Eiffel University – France
david.souriou@fi-ndt.fr

Xavier Dérobert
GERS/GeoEND, Nantes
Gustave Eiffel University – France
xavier.derobert@univ-eiffel.fr

*Abstract*— Among the commonly used non-destructive techniques, the Ground Penetrating Radar (GPR) is one of the most widely adopted today for assessing pavement conditions in France. However, conventional radar systems and their forward processing methods have shown their limitations for the physical and geometrical characterization of very thin layers such as tack coats. However, the use of Machine Learning methods applied to GPR with an inverse approach showed that it was numerically possible to identify the tack coat characteristics despite masking effects due to low time-frequency resolution noted in the raw B-scans. Thus, we propose in this paper to apply the inverse approach based on Machine Learning, already validated in previous works on numerical data, on two experimental cases with different pavement structures. The first case corresponds to a validation on known pavement structures on the Gustave Eiffel University (Nantes, France) with its pavement fatigue carousel and the second case focuses on a new real road in Vendée department (France). In both case studies, the performances of SVM/SVR methods showed the efficiency of supervised learning methods to classify and estimate the emulsion proportioning in the tack coats.

## I. Introduction

To date, it seems difficult to accurately estimate the dielectric and geometric characteristics (complex permittivity, thickness, binder content, etc.) of thin layers such as pavement tack coats directly from *in situ* non-destructive measurements. Some previous research works of [1] have shown that the inversion of GPR (with a Stepped-Frequency Radar) signals using the Full Waveform Inversion (FWI) method allows to describe rigorously the electromagnetic (EM) wave propagation in multi-layered media such as pavements. This work allowed to link qualitatively the equivalent dielectric susceptibility of the tack coat and the bituminous emulsion proportioning (Fig. 1).

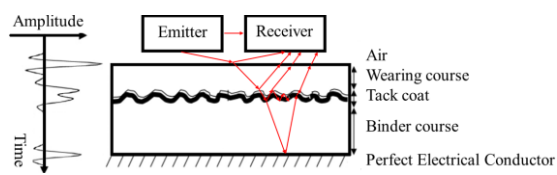
Fig. 1. Modeling scheme of EM wave propagation in multi-layered pavement structure

Thus, the GPR measurements and their inversion results make it possible to identify a monotonous evolution of the estimated dielectric susceptibility of the interfacial transition zone as a function of the controlled bituminous emulsion quantity. However, these works have faced the selectivity complex problem of the analytical model where different geometrical and physical characteristics presented a similar impact on the dielectric response of the multi-layered medium. This is the main reason why the presented research has been focused on processing GPR data using Machine Learning methods. Indeed, previous studies have made it possible to successfully combine SVM/SVR-type Machine Learning methods with GPR techniques for many applications such as the detection and localization of utility networks, the thickness estimation of railway layers, and more generally the estimation of thicknesses (time delay for non-dispersive media) and dielectric permittivities of civil engineering media [2]. These approaches have also been developed in order to extract characteristics related to structural anomalies such as debondings in pavements [3]. The use of this methodology that consists in inverse processing of GPR data by SVM/SVR methods, allowed us to classify the bituminous emulsion proportioning (tack coat) between wearing and binder layers. This was done firstly on synthetic pavement structures modeled in gprMax [4 – 5], then on controlled experimental pavements which are the subject of this paper.

In this paper, the first part (Section II) introduce a brief reminder of the numerical parametric study used to validate the SVM/SVR approach on synthetic signals. The second part (Section III) presents a first experimental campaign carried out on the controlled test platform of the Gustave Eiffel University (Nantes/France). It is composed of different pavement structure sections, with and without tack coat, in a full-scale fatigue carousel that apply a very large range of loadings and degradation stresses. The second experiment studied in this paper corresponds to a real pavement structure in the Vendée department (France) where three different bituminous emulsion proportioning have been performed exclusively for this research work and its tack coat problematic. Finally, the results of the SVM/SVR methods applied with an inverse approach to the GPR signals issued from the two experimental sites are presented and then discussed (Section IV).

## II. METHODS : PRINCIPLE AND NUMERICAL APPROACH

In this part, a reminder of the numerical validation results using SVM/SVR methods on an example of synthetic GPR database is done. This study allowed to classify the features extracted from raw GPR signals according to the presence (or not) of a tack coat following two classes using the Two-Class SVM (TCSVM), four classes (absence, under proportioning, correct proportioning and over proportioning) using the Multi-Class SVM (MCSVM) and with a quantitative estimation of the emulsion quantity using the Support Vector Regression (SVR). The study of these three variants of the SVM method allowed to optimize the data processing protocol for each of the experiments presented in this paper.

Various two-layer structures were modeled using gprMax 3D with different geometries and physical parameters (materials, layers and antennas) in order to validate the ability of GPR wave propagation to detect the presence of a millimetric thin layer such as tack coat. Thus, this numerical approach was used to model realistic pavement structures, 50 meters long and 5 meters wide scanning the whole surface with 0.25 meters steps [5]. By the use of a bivariate normal distribution, various tack coat thicknesses were implemented, defining as well the ground truth. The acquired data were processed using supervised learning methods such as SVM/SVR. Thus, an EM mapping of the simulated pavements have been performed with a ground-coupled antenna at the center frequency fc = 2.6 GHz (Fig. 2).

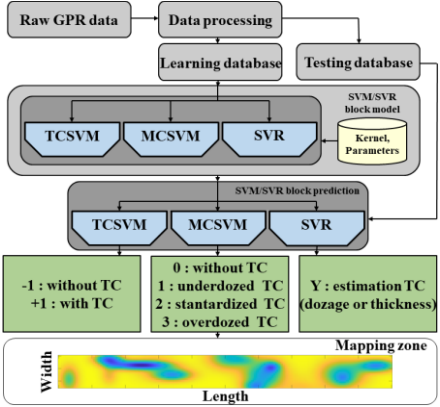

Fig. 2. Diagram describing GPR data processing using SVM/SVR methods to detect, classify and estimate the tack coat thicknesses

The objective of SVM is to find a hyperplane in a N-dimensional space (with N representing the number of A-scan features) allowing data separation into two or more classes. The goal is to maximize the decision limit by choosing a margin that will represent the maximized distance between the variables and the separation vector. The optimization is therefore done by the minimization of the cost function as followed:

$$\begin{cases} \min_{\mathbf{w},b,\zeta} \quad \frac{1}{2}\,||\mathbf{w}||^2 + C\sum_{i=1}^{N}\zeta_i \\ y_i(\mathbf{w}^T\mathbf{x}_i + b) \geq 1 - \zeta_i, \\ \text{with} \quad \zeta_i \geq 0, i = 1, 2, \dots N \end{cases} \quad (1)$$

$\mathbf{x}_i \in \mathbb{R}^N$ is the set of training vectors where $\mathbf{x}_i$ represents the $i$-th A-scan with $i = [1, 2, \dots, N]$ and $N$ is the number of A-scans. $C$ represents the regularization variable and allows finding a compromise between the width of the margin and the false classified points by penalizing them, $y_i \in \{-1, +1\}$ is the result of the classification for each A-scan, $\mathbf{w}$ is the normal vector, $\zeta_i$ is the so-called slack variable (which is therefore a degree of freedom allowing certain training points to be within the margin) and $b$ the bias.

This numerical study allowed to conclude that it was possible to distinguish various emulsion proportioning of a tack coat applied between the first two pavement layers as well as the zones presenting anomalies (Fig. 3). The *in-lab* experimental campaigns allowed the optimization of the studied methods for a use on real civil engineering structures with a parameter setting relied to the characteristics of the prospected zones. The TCSVM method was then used for experimentation on the fatigue carousel platform since it represents pavement structures with, for each of the two sections, two distinct classes of emulsion quantity used for the tack coat proportioning. In addition, the MCSVM/SVR methods were applied to the aforementioned real site in the Vendée departement (France), since it represents a pavement structure with three emulsion quantities used for the tack coat proportioning.

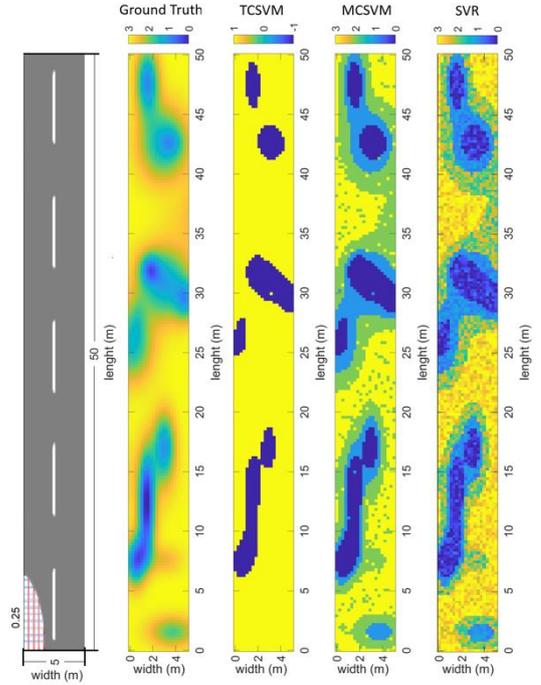

Fig. 3. a) Example of a two-layer pavement structure modeled using gprMax 3D with meshing nodes corresponding to antenna positions, b) Example of the reference mapping representing the spatial variation of the tack coat thicknesses on the whole structure, c) Example of TCSVM results for a binary classification issued from A-scans feature processing, d) Example of MCSVM results for thickness classification issued from A-scans feature processing, e) Example of SVR results for thickness classification issued from A-scans feature processing

## III. EXPERIMENTAL SETUP

### A. Fatigue carousel platform

The fatigue carousel of the Gustave Eiffel University in Nantes (France) is a device that has been used for 40 years for full-scale experiments. Its objective is to carry out accelerated degradation and aging tests on real pavement structures under variable loads (65 kN on single- or double-wheel shaft, tandem or tridem) and at variable speeds (up to

100 km/h). With a total circular length of 120 meters by 6 meters wide, the device allows to reproduce road traffic and therefore the structural aging of a road (Fig. 4).

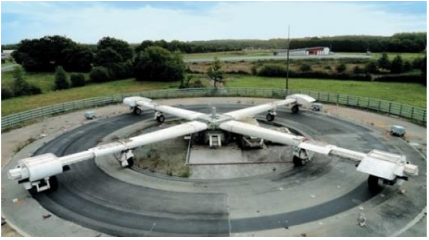

Fig. 4. Image of fatigue carousel platform in Gustave Eiffel University (Nantes/France)

Within the framework of this study and more generally of the Binary project (ANR – French Research Agency), a pavement structure 60 meters long and 3.5 meters wide was performed. This is made up of 5 sections with variable geometries. Sections S1a, S2b and S2c have a tack coat proportioned with 300 g/m² while sections S1b and S2a do not contain any tack coat (Fig. 5) [6].

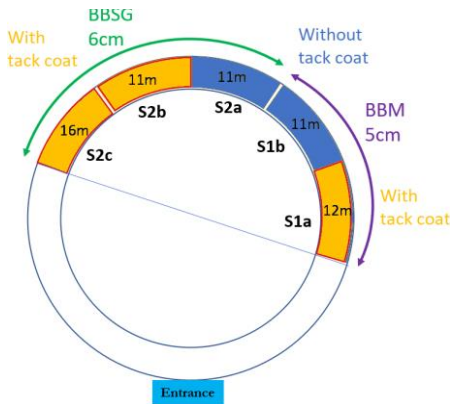

Fig. 5. Diagram representing the different sections made on the fatigue carousel structure

Auscultation tests were carried out with various devices over several months: a GSSI impulse GPR (350 MHz, 900 MHz, 1.5 GHz and 2.6 GHz antennas), a home-made stepped frequency radar (UWBepsilon system) with ultra-wideband air-coupled antennas (400 MHz – 12 GHz) developed by the start-up FI-NDT (spin-off of Gustave Eiffel University) and a 3D-radar system (Kontur – LAMES Gustave Eiffel University Device) with DXG1820 ground-coupled antennas (200 MHz – 3 GHz). In this paper, we choose to present only the results obtained using the commercial GSSI system with 2.6 GHz antennas. The measurements were taken along 6 profiles (4 longitudinal profiles: P1 at 0.8 m from the inside edge, P2 at 1.5 m, P3 at 1.9 m and P4 at 2.65 m; 2 transverse profiles: P5 at 3 m from the start of each section and P6 at 3 m from the end). A spatial sampling of 50 scans/m was used with 2048 sample/scan, this choice was motivated by the results of the numerical parametric study regarding the used SVM/SVR methods.

From the direct data processing (visual analysis of the B-Scans and after applying some basic processing), we clearly identify through the blue marks in the Fig. 6 the change in thickness of the wearing courses between structures BBM (Thin bituminous concrete *in french : Béton Bitumineux Mince*) and BBSG (Semi-coarse HMA *in french : Béton Bitumineux Semi Grenu*) at about 23 meters. On the other hand, it is impossible to visually distinguish the presence or not of the tack coat at 12 and 34 meters. In this configuration, we applied the supervised learning method TCSVM, for which the results will be discussed in part IV, to separate the auscultated structures into two classes (presence or not of the tack coat).

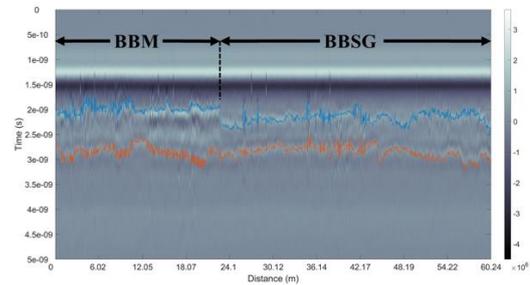

Fig. 6. Example of a B-Scan representing the longitudinal profile P3 with the bottom of the wearing course in blue and the bottom of the binder course in orange

## B. Experimental road in Vendée department (France)

For the second experiment, a new section of real road was adapted for the study in the French department of Vendée. For the research purposes, the manager applied a tack coat with three different emulsion quantities over a distance of 120 meters. Thus, 450 g/m² was applied over the first 40 meters followed by 250 g/m² over the same distance and finally 300 g/m² for the rest of the site. A single geometry was implemented, namely a surface layer thickness BBSG=5 cm over the entire roadway. For each section, the measurements have been done along three longitudinal profiles (P1 at 1.2 m from the right edge, P2 at 2.5 m and P3 at 3.8 m) and 2 transverse profiles per section using the impulse GPR (GSSI with 2.6 GHz ground-coupled antennas). As for the experiment on the fatigue carousel platform, it is possible to identify the bottom of the wearing course from the direct analysis of the B-Scans (Fig. 7). It is also impossible to distinguish visually the different bituminous emulsion quantities at 40 and 80 m. Unlike the previous experiment, the GPR data processing will be performed in this case using the MCSVM method.

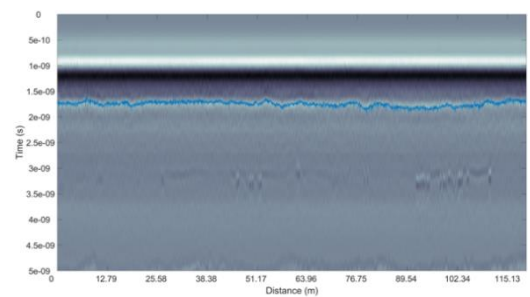

Fig. 7. Example of a B-Scan representing the longitudinal profile P3 with the bottom of the wearing course in blue

## IV. RESULTS AND DISCUSSIONS

The two experiments showed that an identification of the tack coat layer was impossible by a forward approach, which implies the use of supervised learning methods such as SVM/SVR. The fatigue carousel configurations allowed us to establish a TCSVM binary classification (on samples that have not been used for training) in the form of an EM mapping of the studied zones and to accurately identify the spatial distribution of tack coat characteristics (Fig. 8). It should be noted the presence of transition zones at the intersections in addition to overlapping phenomena which generate a systemic error on the optimization process for the hyper-parameter estimation and therefore "outliers" in the results of the classification. The DSC (Dice SCore) classification performance indicator of nearly 93% demonstrates the ability of the TCSVM and its associated Kernel to estimate in a very satisfying way the positioning of healthy areas and areas where the tack coats are presenting defects (Fig. 9).

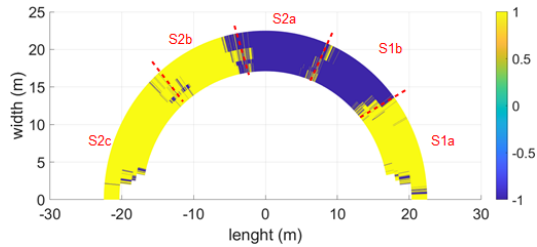

Fig. 8. Classification mapping based on TCSVM applied to GPR data, yellow parts corresponding to zones with tack coat and blue parts corresponding to zones without tack coat

Fig. 9. TCSVM classification confusion matrix of the tested database corresponding to the hole fatigue carousel structure

Concerning the experimentation on the real pavement structure in Vendée department, the training and the tests were carried out on a database representing three different emulsion quantities used to perform the tack coats. Therefore, the use of MCSVM classification and/or SVR regression for EM mapping is more suitable than a simple binary classification using TCSVM method. In the example presented (Fig. 10), the macro DSC performance indicator of classification by MCSVM is greater than 91% (Fig. 11) with errors (RMSE) less than 43. This confirms the ability of the used Machine Learning method to reliably estimate the positioning of healthy and degraded zones in what concern tack coats in the hole real structure (Fig. 10).

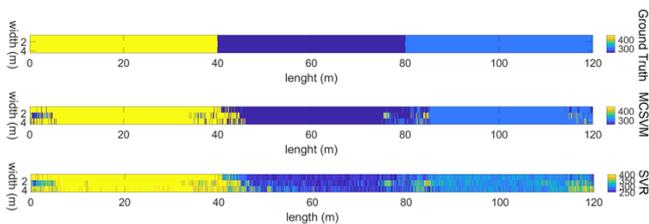

Fig. 10. a) Ground truth distinguishing 3 zones, 450 g/m² in yellow, 250 g/m² in dark blue and 350 g/m² in light blue b) Classification mapping based on MCSVM applied to GPR data c) Regression mapping (estimated values) based on SVR applied to GPR data

## V. CONCLUSION

This research work allowed to successfully combine Machine Learning methods SVM/SVR with GPR techniques for the pavement tack coat characterization. The study briefly presented in this paper has confirmed experimentally (on real structures) the inversion results obtained exclusively in previous researches on numerical data.

Fig. 11. MCSVM classification confusion matrix of the tested database corresponding to the hole pavement structure of Vendée department

Two pavement structures were auscultated in this context, the Gustave Eiffel University fatigue carousel and a new pavement structure in Vendée department (France). The first site with known and controlled formulation with two bituminous emulsion quantities used for the tack coat (0 and 300 g/m²) and the second with three different emulsion quantities (250, 300 and 450 g/m²). After an optimization of the choice of the corresponding method according to the nature of the targeted classes (TCSVM in the first case and MCSVM/SVR in the second), the study confirms with a quantitative aspect the results of the literature, namely the sensitivity of the EM waves to the dielectric and geometric characteristics of the tack coat. The SVM/SVR performances show the effectiveness of supervised learning methods for classifying and estimating emulsion quantities from GPR measurements.

ACKNOWLEDGMENT

The authors would like to thank the French National Research Agency (ANR) which, through the BINARY project, allowed us to conduct this study on the fatigue carousel. They also thank Dominique Pineau for his help in carrying out the experiments in Vendée site. Finally, the Pays-de-la-Loire region (France) is also thanked for its contribution through the DEMOFIT project in the acquisition of 3D-Radar device that was also used in this research work.